%% file: main.tex
\documentclass[10pt,twocolumn,letterpaper]{article}

\usepackage{iccv}              


\definecolor{iccvblue}{rgb}{0.21,0.49,0.74}
\usepackage[pagebackref,breaklinks,colorlinks,allcolors=iccvblue]{hyperref}

\def\paperID{3032} 
\def\confName{ICCV}
\def\confYear{2025}

\usepackage{algorithm}
\usepackage{algorithmic}

\usepackage{color}
\usepackage{booktabs}
\usepackage{amsfonts}
\usepackage{xcolor}
\usepackage{amsmath}
\usepackage{newfloat}
\usepackage{listings}
\usepackage{multirow}
\usepackage{array}
\usepackage{arydshln}
\usepackage{makecell}
\usepackage{amssymb}
\title{EE-MLLM: A Data-Efficient and Compute-Efficient Multimodal Large Language Model}

\author{
    Feipeng Ma$^{1,2}$\thanks{This work was performed during Feipeng Ma's internship at WeChat, Tencent Inc.} 
    \ Yizhou Zhou\textsuperscript{2}
    \ Zheyu Zhang\textsuperscript{1}
    \ Shilin Yan\textsuperscript{3}
    \ Hebei Li\textsuperscript{1}
    \ Zilong He\textsuperscript{1} \\
    \ Siying Wu\textsuperscript{4}
    \ Fengyun Rao\textsuperscript{2}
    \ Yueyi Zhang\textsuperscript{1\textdagger} 
    \ Xiaoyan Sun\textsuperscript{1,4}\thanks{Corresponding authors.} \\
    \textsuperscript{1}University of Science and Technology of China \ \textsuperscript{2}WeChat, Tencent Inc. \
    \textsuperscript{3}Fudan University \\
    \textsuperscript{4}Institute of Artificial Intelligence, Hefei Comprehensive National Science Center \\
    {\tt\small mafp@mail.ustc.edu.cn \ harryizzhou@tencent.com}\\
    \tt\small{\{zhyuey,sunxiaoyan\}@ustc.edu.cn}
}

\begin{document}
\maketitle
\input{Sections/0_abstract}
\input{Sections/1_intro}

\input{Sections/2_related_work}

\input{Sections/3_method}
\input{Sections/4_experiment}

\input{Sections/5_conclusion}

{
    \small
    \bibliographystyle{ieeenat_fullname}
    \bibliography{main}
}
\input{Sections/6_appendix}


\end{document}

%% file: Sections/0_abstract.tex
\begin{abstract}
Recent advancements in Multimodal Large Language Models (MLLMs) have demonstrated satisfactory performance across various vision-language tasks. Current approaches for vision and language interaction fall into two categories: self-attention-based and cross-attention-based methods. However, both approaches present inherent limitations, forcing a trade-off between data and computational efficiency. To address this issue, we introduce the Data-\textbf{E}fficient and Compute-\textbf{E}fficient \textbf{MLLM} (\textbf{EE-MLLM}). Specifically, we modify the original self-attention mechanism in MLLM to a composite attention mechanism. This mechanism has two key characteristics: 1) eliminating the computational overhead of self-attention among visual tokens to achieve \textbf{compute efficiency}, and 2) reusing the weights from each layer of LLM to facilitate effective vision-language modality alignment for \textbf{data efficiency}. As a result, EE-MLLM significantly outperforms Flamingo with limited training data, and reduces the prefilling time to 79 ms on an H800 GPU, compared to LLaVA's 277 ms. To further investigate the efficiency of EE-MLLM, we present a training-free variant named EE-MLLM-F, which reduces the computation cost of self-attention-based method without additional training. Experimental results demonstrate the effectiveness of EE-MLLM across a range of benchmarks, including general-purpose datasets like MMBench and SeedBench, as well as fine-grained tasks such as TextVQA and DocVQA. 
\end{abstract}

%% file: Sections/1_intro.tex
\begin{figure*}[ht]
    \centering
    \includegraphics[width=0.95\linewidth]{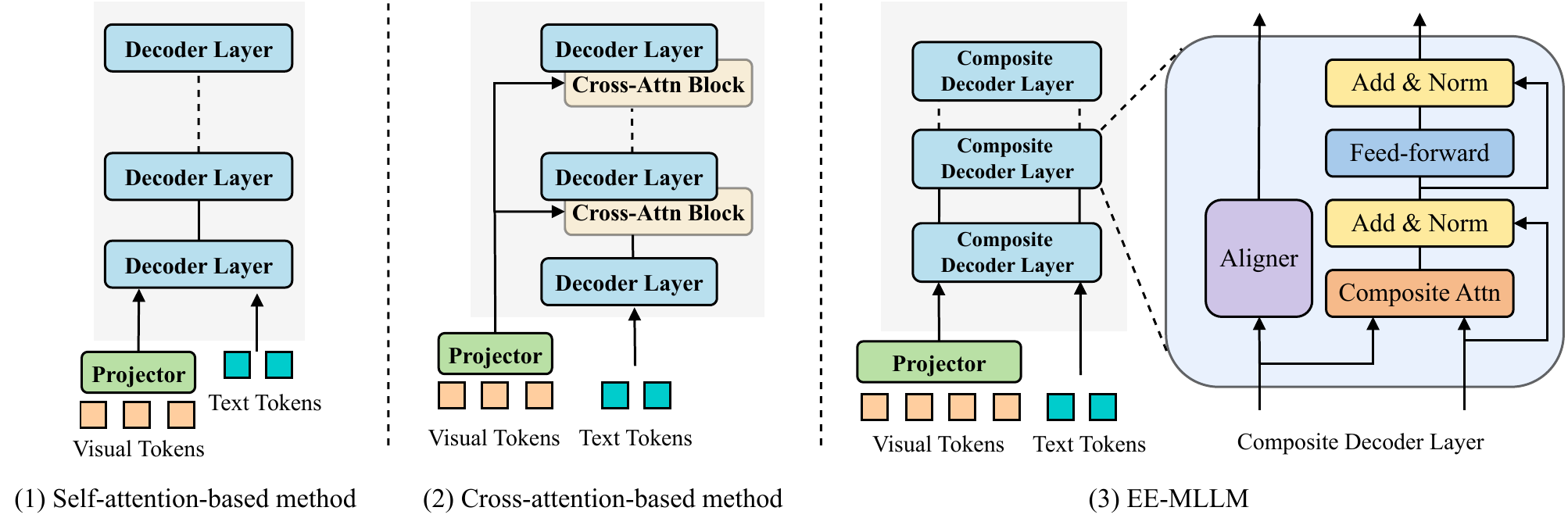}
    \caption{Architecture comparisons between self-attention-based method, cross-attention-based method and our EE-MLLM. (1) The self-attention-based mechanism utilizes a projector to align visual tokens with text tokens, subsequently concatenating these tokens as input for the LLM. (2) The cross-attention-based mechanism integrates additional cross-attention blocks into the decoder layers. (3) EE-MLLM introduces a composite attention mechanism to eliminate the computational overhead of self-attention within visual tokens and reuse the weights as aligners on each layer of LLM to facilitate modality alignment.}
    \label{fig:overview}
\end{figure*}
\section{Introduction}
Multimodal Large Language Models (MLLMs) have gained significant attention due to their remarkable performance on diverse vision-language tasks, including image captioning, Visual Question Answering (VQA) and so on. 
Currently, mainstream MLLMs can be categorized into self-attention-based and cross-attention-based methods. However, these two predominant approaches entail a trade-off between compute and data efficiency. 
In general, cross-attention-based methods excel in computational efficiency but require substantial training data for effective modality alignment, thus exhibiting low data efficiency. Conversely, self-attention-based methods have high data efficiency, which can achieve modality alignment with limited data, but often suffer from high computational complexity, resulting in reduced computational efficiency.

Specifically, self-attention-based methods project visual tokens into LLM's input space with a simple MLP and concatenate them with text tokens, as illustrated in Fig.\ref{fig:overview} (1). 
Their data efficiency stems from: 1) Few learnable parameters in the alignment module, and 2) 
When visual tokens are aligned to the input space via a projector during input, these remain consistent with the processing space of each LLM's decoder layer.
Therefore, LLaVA achieves strong performance with only 558k pre-training and 665k instruction-tuning samples~\cite{liu2023llava}, in other words, it does not require a large amount of data for modality alignment.
Cross-attention-based methods insert cross-attention layers between existing LLM layers. As shown in Fig.\ref{fig:overview} (2), the keys and values in these layers are derived from visual tokens, while the queries come from text tokens. These methods introduce a large number of learnable parameters, which require substantial training data, leading to data inefficiency.
For example, Flamingo is pre-trained on 1.8 billion image-text pairs~\cite{alayrac2022flamingo, jia2021scaling_align} and interleaved data. 
Thus, this data inefficiency arises from two factors: 1) Learnable parameters from additional cross-attention layers, and 2) The training procedure has to focus on the modality alignment between vision and language on different layers of LLM, posing a huge optimization complexity for modality alignment.

Regarding compute efficiency, self-attention methods concatenate visual and text tokens, which increases sequence length from $T$ to $V+T$, raising computational complexity of LLM from $O(T^2)$ to $O((V+T)^2)$. In contrast, cross-attention methods can achieve high compute efficiency because the input sequence length remains fixed at $T$ and does not increase with the number of visual tokens.

By reviewing existing methods, we conclude that the key points to ensure compute efficiency and data efficiency are 
1) Avoiding an increase in the length of input sequence for LLM, 2) The modality alignment module should have fewer parameters, and the alignment needs to be effective.
Taking two summary viewpoints into account, we present a data-\textbf{E}fficient and compute-\textbf{E}fficient \textbf{M}ultimodal \textbf{L}arge \textbf{L}anguage \textbf{M}odel (\textbf{EE-MLLM}). 
Specifically, we propose a composite attention mechanism to replace the original self-attention mechanism in LLM. The composite attention mechanism has two features: 1) Eliminating the computational overhead of self-attention within visual tokens to achieve \textbf{compute efficiency}. 
Similar to FastV~\cite{chen2024image_fastv}, we observe that the self-attention within visual tokens is redundant. This redundancy stems from: a) the interaction between visual tokens is already well-processed by the vision encoder, and b) LLM can facilitate the implicit interaction among visual tokens via the information aggregation property of LLM~\cite{wang2023label}.
Therefore, we eliminate self-attention among visual tokens and maintain the length of the input sequence for the LLM to be the same as that of text tokens, thereby facilitating computational efficiency.
2) Instead of introducing additional modules like cross-attention-based methods, we reuse the weights on each layer of LLM to facilitate effective modality alignment between vision and language for \textbf{data efficiency}. 
As a result, without introducing additional parameters, visual tokens aligned in the input space of LLM can naturally facilitate the alignment between vision and language in each layer of LLM. This means the training procedure does not need to focus on the modality alignment within the LLM, which enables data efficiency. With the proposed composite attention, EE-MLLM can achieve data-efficient and compute-efficient at the same time.
We also investigate a training-free variant of EE-MLLM where we directly apply the composite attention mechanism to select layers of LLaVA, thereby reducing computational cost without requiring additional training. Notably, this approach maintains comparable performance, demonstrating computational efficiency without sacrificing accuracy.

We evaluate EE-MLLM on both general benchmarks and fine-grained benchmarks and our model demonstrates promising results across all these evaluations while maintaining high computational efficiency during inference.
For a $980 \times 980$ input image, EE-MLLM requires only $76\%$ of the FLOPs needed by self-attention methods. 
Moreover, in real-time scenarios that require high response speed but fewer output tokens, EE-MLLM can significantly reduce the prefilling time. On a single NVIDIA H800 GPU, EE-MLLM takes only 79 ms for prefilling with one input image, whereas LLaVA consumes 277 ms.

\begin{figure*}[t]
    \centering    \includegraphics[width=0.73\linewidth]{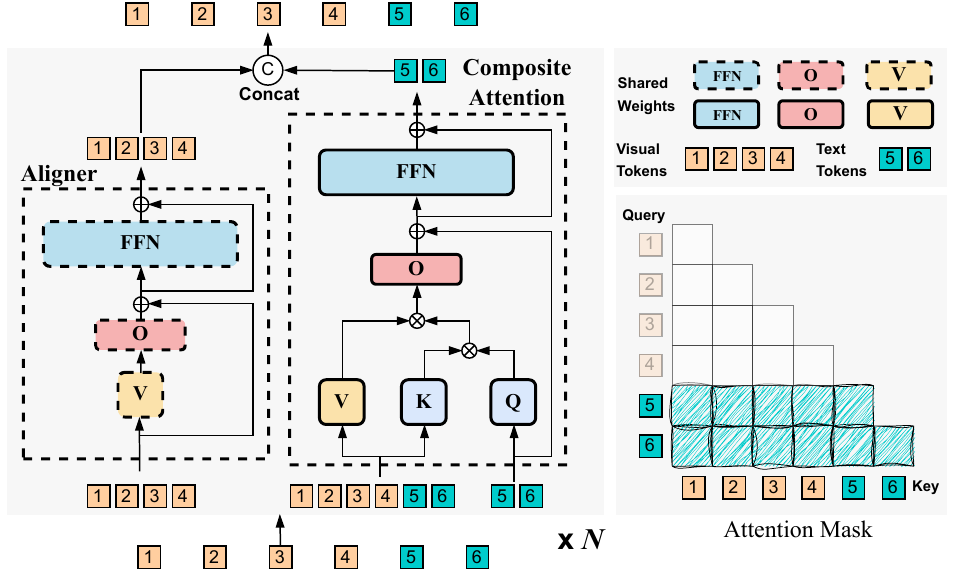}
    \caption{Our composite attention mechanism consists of the composite attention module and the aligner. 
    For the aligner, visual tokens are aligned to the feature space of each layer of LLM by the aligner alone; for the composite attention module, the concatenation of visual and text tokens are used as keys and values, and the text tokens are used as queries for attention, thus eliminates the self-attention within visual tokens.
    }
    \label{fig:method}
\end{figure*}
Our main contributions are summarized as follows:
\begin{enumerate}
    \item We propose composite attention mechanism for EE-MLLM, which enhances both data and compute efficiency by eliminating redundant attention calculation among visual tokens and reusing existing LLM weights for modality alignment.
    \item We also introduce a training-free variant EE-MLLM-F, which adapts the composite attention mechanism to existing self-attention-based models, reducing computational costs without requiring additional training while preserving performance.
    \item Experimental results demonstrate that EE-MLLM achieves superior performance across various benchmarks, while requiring only 76\% of the FLOPs compared to self-attention methods and significantly reducing the prefilling time from 277 ms to 79 ms.
\end{enumerate}

%% file: Sections/2_related_work.tex
\section{Related Work}
\noindent\textbf{Self-attention-based methods.}
Self-attention-based methods~\cite{dai2024instructblip,zhu2023minigpt,chen2023minigpt4_v2,bai2023qwen,chen2023sharegpt4v,fang2023instructseq,lu2024deepseek} employ a projector to align visual features with the input space of LLM. LLaVA~\cite{liu2023llava,liu2023llava_1p5} uses a simple MLP layer to connect the vision encoder of CLIP~\cite{openaiclip} with Vicuna~\cite{zheng2023vicuna}. 
Considering the scenarios that require fine-grained information, such as document recognition and form understanding, recent work extends MLLM to support high-resolution inputs, resulting in long input sequences with numerous visual tokens.
LLaVA-Next~\cite{liu2024llavanext} proposes dividing the image into small patches for extracting local features and concatenating the downsampled image feature as global context. Monkey~\cite{li2023monkey} also partitions high-resolution images into small patches, processing each patch independently with a trainable resampler.
Another line of research focuses on compressing the number of visual tokens to reduce computational cost. Deco~\cite{yao2024deco} and HoneyBee~\cite{cha2024honeybee} carefully design projectors to reduce the number of visual tokens with minimal information loss. However, these methods still compromise fine-grained capabilities when reducing the number of visual tokens.
Our EE-MLLM, a cross-attention-based architecture, achieves state-of-the-art performance while maintaining computational efficiency.

\noindent\textbf{Cross-attention-based methods.}
Cross-attention-based methods~\cite{laurencon2023obelics, alayrac2022flamingo,awadalla2023openflamingo,chen2024evlm} typically incorporate additional cross-attention modules into the decoder layers of LLM to integrate visual information without extending the length of the text token sequence.
Flamingo~\cite{alayrac2022flamingo} introduces a novel gated xattn-dense block, incorporating a tanh-gating mechanism for training stability. Flamingo is trained on large-scale interleaved image-text data and supports in-context few-shot learning.
EVLM~\cite{chen2024evlm} enhances Flamingo by: (1) utilizing hierarchical ViT features without a resampler, (2) replacing the image token with a set of learnable tokens in the text input, and (3) introducing MoE to boost performance.
Although these cross-attention-based methods are computationally efficient, the additional cross-attention blocks lead to a large number of trainable parameters and a heavy reliance on pre-training data. Our EE-MLLM avoids introducing any extra trainable parameters and repurposes the weights of LLM to map visual tokens into the LLM feature space, thus achieving promising performance with a small amount of training data.

%% file: Sections/3_method.tex
\section{Method}
\subsection{Model Architecture}
EE-MLLM is composed of three main components: a visual encoder, a projector made up of two-layer MLPs, and an LLM with composite decoder layers. The composite decoder layer contains an aligner and a composite attention module, designed to ensure data and compute efficiency.

\subsubsection{Composite Attention Module}
The original self-attention mechanism builds correlations in three aspects: intra-interaction among visual tokens, intra-interaction among text tokens, and inter-interaction among visual tokens and text tokens.
In decoder layers, we observe that the process with visual tokens is redundant for two reasons: 1) The interaction among visual tokens is already well-learned in the vision encoder, and 2) the visual information has been aggregated into the text tokens through the inter-interaction among text tokens and visual tokens~\cite{wang2023label}. Therefore, we eliminate intra-interaction among visual tokens and obtain the composite attention module.

We assume that visual tokens $I\in \mathbb{R}^{k\times h}$ and text tokens $T\in \mathbb{R}^{n\times h}$, where $k$ and $n$ are the lengths of visual tokens and text tokens, respectively. And $h$ denotes the dimension of hidden states.
The original self-attention can be formulated as follows:
\begin{equation}
    \textrm{Attention}(X^q, X^k, X^v)=\textrm{softmax}(\frac{X^q{X^k}^T}{\sqrt{h}})X^v ,
\end{equation}
where $X^q = [I;T] W_q$, $X^k = [I;T] W_k$, $X^v = [I;T] W_v$, $W_q$, $W_k$ and $W_v$ are the weights of the self-attention module. The symbol $[;]$ denotes the concatenation operation of two variables in the sequence dimension.

As illustrated in Fig.~\ref{fig:method}, our composite attention module uses text tokens as queries and the concatenation of visual tokens and text tokens as keys and values for cross-attention. Specifically, $X^q = T W_q$, $X^k = [I;T] W_k$, and $X^v = [I;T] W_v$. The length of the output sequence will be $n$, which is the same as the length of text tokens $T$.
The attention mask for self-attention is a lower triangular matrix, whereas the attention mask for the composite attention module is a trapezoidal matrix. This implies that text tokens will attend to all previous tokens, including visual tokens, while intra-interaction among visual tokens is eliminated.

\subsubsection{Aligner}
We introduce the aligner, which leverages the existing weights at each layer of LLM to enhance modality alignment between vision and language without the need for additional modules such as cross-attention-based methods. By reusing these weights, visual tokens aligned in the input space of the LLM can naturally facilitate modality alignment at each layer. This approach eliminates the necessity for the training procedure to concentrate on modality alignment within the LLM, thereby promoting data efficiency. 

As depicted in Fig.~\ref{fig:method}, in each layer, text tokens are mapped by the value matrix ($W_v$), output matrix ($W_o$), and FFN. Hence, we make visual tokens follow the same mapping relationship in the aligner.
Specifically, the aligner comprises $W_v$, $W_o$, and FFN, which share weights with the corresponding LLM modules. The residual connection is also implemented in the aligner.
The aligner can be formulated as follows:
\begin{align}
    O = \mathrm{FFN}(I W_v W_o + I) + I W_v W_o + I
\end{align}
This part of the decoder layer skips attention calculation.

\subsection{Training-Free Adaptation of EE-MLLM}
Since our composite attention mechanism does not introduce additional parameters for the interaction between vision and language, which reuses the weights of each LLM layer for modality alignment, we can directly apply this method to self-attention-based models without requiring additional training. 
Specifically, given a well-trained MLLM with $L$ layers, we can replace the self-attention blocks in selected layers $\mathcal{L}_{\text{replace}} \subseteq \{1,2,...,L\}$ with our composite attention.
For each layer $l \in \mathcal{L}_{\text{replace}}$, the original self-attention is substituted with our composite attention, and the weights of $W_v$, $W_o$, and FFN are reused to align the visual tokens. 

Inspired by FastV~\cite{chen2024image_fastv}, which demonstrates the attention of visual tokens becomes sparse in deep layers, we maintain the self-attention mechanism in the early layers of LLM while replacing the deeper layers with our composite attention. 
With this adaption, we can reduce the computation cost of the self-attention-based MLLM in deep layers, while maintaining strong performance.

\begin{table*}[t]
    \centering
    \resizebox{\linewidth}{!}{
    \begin{tabular}{l c c c c c c c c c}
    \toprule
    \rowcolor{gray!10} \textbf{Model} & \textbf{LLM} & \textbf{MMB-T} & \textbf{MME} & \textbf{ScienceQA} & \textbf{HallB} & \textbf{MMMU} & \textbf{CCBench} & \textbf{SeedBench} & \textbf{BLINK} \\
    \midrule
    InstructBLIP & Vicuna-7B & 36.0 & 1137.1 & 54.7 & 31.2 & 30.6 & 12.7 & 44.5 & --- \\
    MiniGPT-4-v1 & Vicuna-7B & 12.2 & 770.6 & 39.0 & 31.9 & 23.6 & 1.8 & 31.6 & --- \\
    MiniGPT-4-v2 & Llama2-7B & 24.3 & 708.4 & 54.1 & 30.0 & 25.0 & 1.4 & 29.4 & --- \\
    Idefics-instruct & Llama-7B & 48.2 & 942.0 & 51.6 & 27.3 & 18.4 & 7.8 & 45.0 & 38.3 \\
    OpenFlamingo v2 & MPT-7B & 6.6 & 535.0 & 45.7 & 29.4 & 28.2 & 6.3 & 28.8 & --- \\
    Qwen-VL & Qwen-7B & 38.2 & 334.1 & 57.7 & 29.9 & 29.6 & 6.1 & 52.5 & 27.9 \\
    Qwen-VL-Chat & Qwen-7B & 60.6 & 1467.8 & 65.5 & 36.8 & 37.0 & 41.2 & 64.8 & 28.2 \\
    VLoRA & Vicuna-7B & 63.4 & 1311.3 & 66.4 & 26.4 & 36.0 & 28.6 & --- & --- \\
    ShareGPT4V & Vicuna-7B & 64.6 & \textbf{1561.4} & 68.2 & 28.6 & 37.2 & 30.8 & 69.3 & 40.9 \\
    LLaVA-v1.5 & Vicuna-7B & 62.3 & 1510.7 & 66.8 & 27.6 & 35.7 & 27.5 & 65.8 & 39.7 \\
    LLaVA-v1.6 & Vicuna-7B & 66.5 & 1475.6 & 68.5 & 27.6 & \textbf{37.6} & 24.3 & 69.6 & 41.6 \\
    \midrule
    \rowcolor{gray!10} EE-MLLM & Vicuna-7B & \textbf{70.4} & 1528.1 & \textbf{77.7} & \textbf{38.6} & 33.4 & \textbf{37.3} & \textbf{70.2} & \textbf{43.2} \\
    \bottomrule
    \end{tabular}
    }
    \caption{Comparison with state-of-the-art methods on general benchmarks, including MMBench, MME, ScienceQA, HallusionBench, MMMU, CCBench, SeedBench, and BLINK. MMB-T indicates we report the results on MMBench-TEST-EN-V11.}
    \label{tab:main_result}
\end{table*}

\begin{table*}[t]
    \centering
    \resizebox{0.8\linewidth}{!}{
    \begin{tabular}{l c c c c c c}
    \toprule
    \rowcolor{gray!10} \textbf{Model} & \textbf{LLM} & \textbf{AI2D} & \textbf{OCRBench} & \textbf{TextVQA} & \textbf{ChartQA}  & \textbf{Seed2 Plus} \\
    \midrule
    InstructBLIP & Vicuna-7B & 40.6 & 27.6 & 33.6 & 10.9  & 29.5 \\
    MiniGPT-4-v1 & Vicuna-7B & 28.4 & 17.2 & --- & ---  & 15.2 \\
    MiniGPT-4-v2 & Llama2-7B & 30.5 & 3.1 & --- & --- & 23.3 \\
    Idefics-instruct & Llama-7B & 42.2 & 25.2 & --- & ---  & 35.4 \\
    OpenFlamingo v2 & MPT-7B & 31.7 & 14.9 & 16.3 & ---  & 28.7 \\
    Qwen-VL & Qwen-7B & 57.7 & 12.7 & 63.1 & 59.0  & 40.1 \\
    Qwen-VL-Chat & Qwen-7B & 63.0 & 48.8 & 60.7 & 49.8  & 40.6 \\
    ShareGPT4V & Vicuna-7B & 58.0 & 37.1 & 51.1 & 21.3 & 46.1 \\
    LLaVA-v1.5 & Vicuna-7B & 55.5 & 31.8 & 45.5 & 17.8  & 41.3 \\
    LLaVA-v1.6 & Vicuna-7B & \textbf{67.0} & 53.2 & 64.4 & 55.4  & 51.6 \\
    \midrule
    \rowcolor{gray!10} EE-MLLM & Vicuna-7B & \underline{65.7} & \textbf{57.4} & \textbf{69.0} & \textbf{68.4} & \textbf{53.5} \\
    \bottomrule
    \end{tabular}
    }
    \caption{Comparison with state-of-the-art methods on fine-grained benchmarks, including AI2D, OCRBench, TextVQA, ChartQA, and SeedBench-2 Plus.}
    \label{tab:finegrained_result}
\end{table*}

\subsection{Analysis of the Computational Overhead}
EE-MLLM reduces computational overhead without introducing extra parameters. This section analyzes the compute efficiency of EE-MLLM compared to LLaVA in terms of FLOPs. We consider an LLM with $d$ blocks, hidden state dimension $h$, intermediate size $m$, $T$ text tokens, and $V$ visual tokens.

The total FLOPs of LLaVA are given by:
{\small
\begin{equation}
\text{FLOPs}_{\text{LLaVA}} = \underbrace{8(V+T)d h^2 + 4(V+T)^2 d h}_{\text{self-attention module}} + \underbrace{6(V+T)d h m}_{\text{FFN module}}.
\end{equation}
}
 
The total FLOPs of EE-MLLM are given by:
{\small
\begin{equation}
\begin{split}
\text{FLOPs}_{\text{EE-MLLM}} 
&= \underbrace{(6V+8T)dh^2 + (2T^2 + 2VT)dh}_{\text{composite attention module}} \\
&\quad + \underbrace{6(V+T)dhm}_{\text{FFN module}}.
\end{split}
\end{equation}
}

For a typical training scenario with a $980\times980$ input image (yielding 4,900 visual tokens) and 256 text tokens, EE-MLLM reduces computational requirements by approximately \textbf{24\%} compared to LLaVA, demonstrating significant efficiency gains.

%% file: Sections/4_experiment.tex
\section{Experiments}
\subsection{Implementation Details}
\noindent \textbf{Model Configurations.}
We employ Vicuna-7b-v1.5~\cite{zheng2023vicuna} as our LLM and SigLIP~\cite{zhai2023sigmoid_siglip} as the vision encoder. Specifically, SigLIP is initialized from Idefics2~\cite{laurenccon2024matters_idefics2}, which supports dynamic resolutions up to 980 $\times$ 980. The projector consists of a two-layer MLP.

\noindent \textbf{Pre-training Configurations.}
During the pre-training stage, we follow LLaVA to adopt blip-558k~\cite{li2022blip} for training one epoch. 
We freeze the LLM and the vision encoder, making only the projector trainable. We adopt AdamW~\cite{loshchilov2018decoupled_adamw} optimizer.
The learning rate is 1e-3, followed by a linear warm-up scheduler and then a cosine decay scheduler. 
The global batch size is 256. 

\noindent \textbf{Fine-tuning Configurations.}
During the fine-tuning stage, we freeze the vision encoder and update the weights of the LLM and projector. The total amount of supervised fine-tuning data is 3 million samples. The sources of fine-tuning data are provided in Appendix~\ref{sec:sft_data}.
We utilize the AdamW optimizer with a global batch size of 128 and a learning rate of 2e-5. The learning rate scheduler is the same as the pre-training stage. 
Training is conducted on 16 H800 GPUs.

\subsection{Evalution Benchmarks}
We conduct our evaluations using the VLMEvalKit~\cite{duan2024vlmevalkit}, and the results of other state-of-the-art models are obtained from the same source.

\noindent \textbf{General Benchmarks.}
For general benchmarks, our evaluation includes MMBench-EN~\cite{MMBench}, MME~\cite{fu2023mme}, ScienceQA~\cite{lu2022learn_scienceqa}, HallusionBench~\cite{guan2023hallusionbench}, MMMU~\cite{yue2023mmmu}, CCBench~\cite{MMBench}, SeedBench~\cite{li2023seed}, and BLINK~\cite{fu2024blink}. Among them, MMBench-EN is a comprehensive multimodal benchmark specifically designed to assess the performance of multimodal large language models (MLLMs). It contains more than 3,000 multiple-choice questions spanning 20 distinct ability categories. We evaluate EE-MLLM on MMBench-EN-V1.1. Further details on the other general benchmarks are provided in Appendix~\ref{sec:general_benchmarks}.

\noindent \textbf{Fine-grained Benchmarks.}
We also compare EE-MLLM with other models on AI2D~\cite{kembhavi2016diagram_ai2d}, OCRBench~\cite{liu2023hidden_ocrbench}, TextVQA~\cite{singh2019textvqa}, ChartQA~\cite{masry2022chartqa}, DocVQA~\cite{mathew2021docvqa}, and SeedBench 2 Plus~\cite{li2024seedbench_plus}. More details of these fine-grained benchmarks are provided in Appendix~\ref{sec:finegrained_benchmarks}.

\begin{figure}[t]
    \centering
    \includegraphics[width=0.95\linewidth]{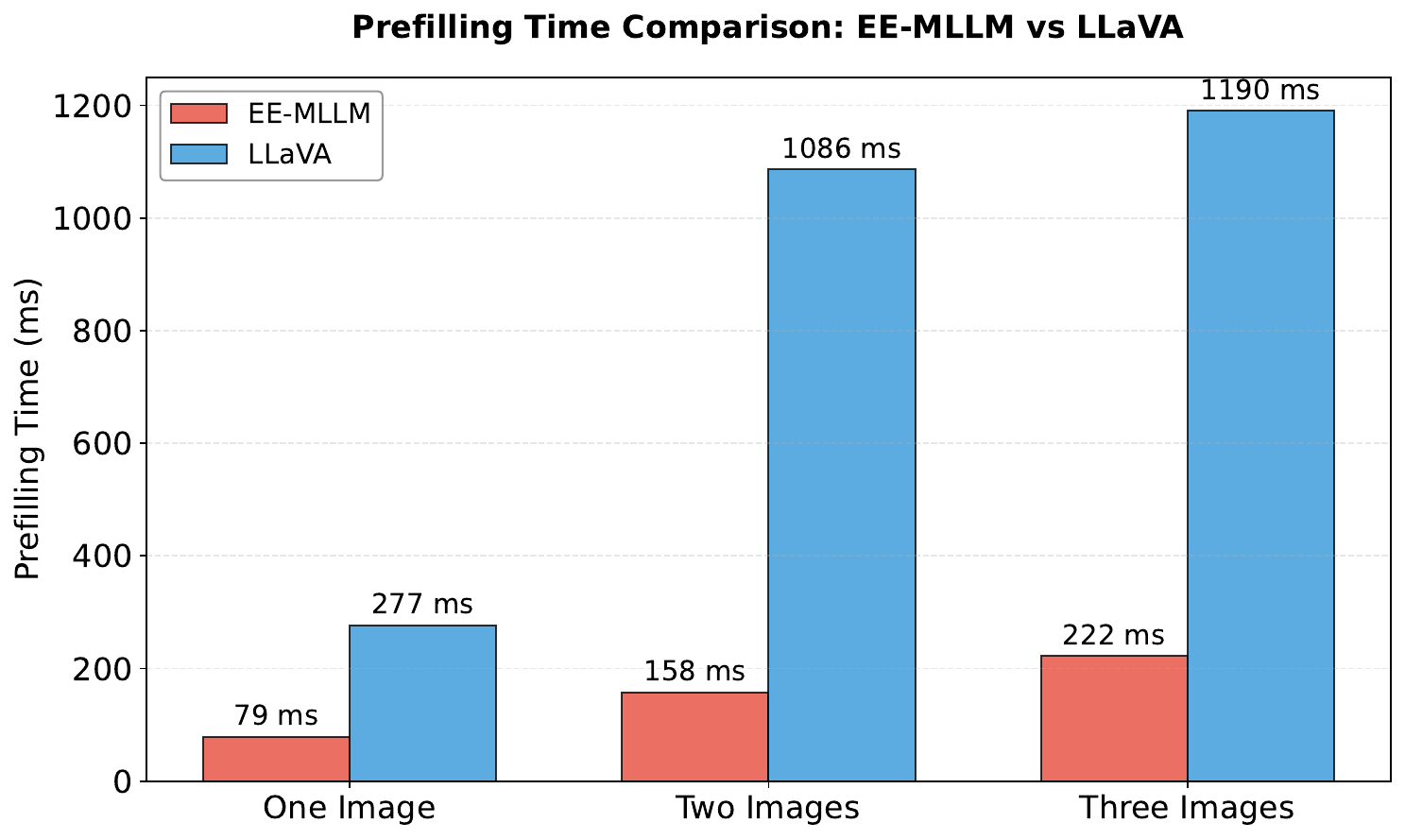}
    \caption{
    Comparison of prefilling time between EE-MLLM and LLaVA. 
    The X-axis represents the number of input images, where each image has a resolution of $980 \times 980$. The Y-axis indicates the prefilling time.
    }
    \label{fig:inference_speed}
\end{figure}

\subsection{Comparisons with State-of-the-art Models}

\noindent \textbf{General Benchmarks.}
In Tab.~\ref{tab:main_result}, we compare EE-MLLM with various  MLLMs equipped similar LLM~\cite{liu2024llavanext, bai2023qwen,ma2024visual_vlora} on general benchmarks. 
These benchmarks evaluate the comprehensive ability of MLLMs, including understanding and perception. 
EE-MLLM achieves comparable performance with state-of-the-art MLLMs in general benchmarks. Specifically, EE-MLLM achieves a score of 70.4 in MMBench and 1528.1 in MME, scores that are notably higer compared to LLaVA-v1.6, which also support high-resolution images input. This suggests that EE-MLLM possesses comprehensive perception and reasoning capabilities. What's more, EE-MLLM also achieves promising performance on CCBench and SeedBench.

\noindent \textbf{Fine-grained Benchmarks.}
In Tab.~\ref{tab:finegrained_result}, we evaluate EE-MLLM on fine-grained benchmarks. These benchmarks demand strong visual perception capabilities from MLLMs, as they require extracting fine-grained information from images to accurately answer questions. 
On TextVQA and ChartQA, EE-MLLM achieves impressive performance, 4.6 higher on TextVQA and 13.0 higher on ChartQA compared with LLaVA-v1.6, which also supports high resolution. In the OCRBench, specifically designed to assess the OCR capabilities of MLLM, EE-MLLM outperforms by 4.2 compared to LLaVA-v1.6. These results demonstrate that despite EE-MLLM significantly reducing the computational overhead associated with visual tokens, it effectively maintains the model's fine-grained capabilities.

\begin{table}[t]
    \centering
    \resizebox{0.95\linewidth}{!}{
    \begin{tabular}{l cccc}
    \toprule
    \textbf{Model} & \textbf{MMB-VAL} & \textbf{SeedBench} & \textbf{ScienceQA} & \textbf{HallB} \\
    \midrule
    LLaVA-v1.5 & 67.3 & 65.1 & 67.0 & 25.7 \\
    EE-MLLM-F & 67.4 & 65.1 & 67.0 & 32.4 \\
    \midrule
    & \textbf{AI2D} & \textbf{OCRBench} & \textbf{TextVQA} & \textbf{ChartQA} \\
    \midrule
    LLaVA-v1.5 & 56.4 & 31.6 & 46.1 & 17.9 \\
    EE-MLLM-F & 56.4 & 30.7 & 45.5 & 17.4 \\
    \bottomrule
    \end{tabular}
    }
    \caption{Performance comparison of LLaVA-v1.5 with our composite attention mechanism across general and fine-grained benchmarks.}
    \label{tab:results_of_training_free}
\end{table}

\subsection{Comparison of Prefilling Time}
Since we eliminate the intra-interactions among visual tokens, 
EE-MLLM can significantly reduce the prefilling time, thus accelerating the generation of the first output token. This efficiency is crucial in real-time scenarios, particularly for tasks that only require a few output tokens. For example, in content moderation systems, MLLMs must rapidly classify social medias as safe or unsafe quickly to prevent harmful content exposure while maintaining platform throughput. Similarly, industrial quality control systems require MLLMs to swiftly provide pass/fail decisions.
We conduct comparisons of prefilling time on a single NVIDIA H800. The resolution of the input image is set to $980\times 980$. We illustrate the time consumed for prefilling by EE-MLLM and LLaVA in Fig.~\ref{fig:inference_speed}.

With one input image, EE-MLLM costs 79 ms for prefilling, while LLaVA costs 277 ms. When the input is extended to two images, the time consumed by LLaVA severely increases to 1086 ms, but the prefilling time of EE-MLLM only rises to 158 ms. We also test the scenario where the number of input images is 3, and the gap between EE-MLLM and LLaVA becomes even larger.

To further analyze the efficiency of EE-MLLM, we compare the overall inference speed and provide a detailed analysis in Appendix~\ref{sec:comp_overall_inference_speed}.

\subsection{Training-free Adaption of EE-MLLM}
In Tab~\ref{tab:results_of_training_free}, we present the results of directly applying our composite attention mechanism to LLaVA-v1.5-7b without additional training. Specifically, we replace the attention mechanism in the last 16 layers of LLaVA-v1.5-7b with our composite attention approach. And EE-MLLM-F performs comparably to the original LLaVA-v1.5 on MMBench, SeedBench, and ScienceQA, and even achieves better results on HallusionBench. For fine-grained benchmarks, we observe only a slight performance drop on OCRBench and TextVQA. These results demonstrate that our composite attention mechanism can directly adapt to self-attention-based methods without the need for training, while maintaining performance without significant degradation.

\begin{table*}[t]
    \centering
    \resizebox{0.95\linewidth}{!}{
    \begin{tabular}{l ccccccccc c}
    \toprule
    \textbf{Model}  & \textbf{MMB-VAL} & \textbf{SeedBench} & \textbf{ScienceQA} & \textbf{HallB} & \textbf{AI2D} & \textbf{OCRBench} & \textbf{TextVQA} & \textbf{ChartQA} & \textbf{Avg.} & \textbf{Ratio} \\
    \midrule
    \rowcolor{gray!10}
    \multicolumn{11}{c}{\textbf{Resolution: $336 \times 336$}} \\
    \midrule
    LLaVA-v1.5  & 67.3 & 65.1 & 67.0 & 25.7 & 56.4 & 31.6 & 46.1 & 17.9 & 47.1 & 100\% \\
    Flamingo  & 60.7 & 53.0 & 66.9 & 28.2 & 52.2 & 13.0 & 27.4 & 12.6 & 39.3 & 81.7\% \\
    EE-MLLM  & 66.8 & 64.1 & 66.7 & 25.2 & 56.2 & 29.1 & 43.3 & 17.3 & 46.1 & 97.9\% \\
    \midrule
    \rowcolor{gray!10}
    \multicolumn{11}{c}{\textbf{Resolution: $672 \times 672$}} \\
    \midrule
    LLaVA-v1.5  & 70.1 & 68.0 & 69.4 & 27.2 & 58.3 & 39.8 & 61.7 & 27.9 & 52.8 & 100\% \\
    Flamingo  & 64.2 & 55.6 & 69.4 & 28.1 & 53.9 & 16.5 & 36.7 & 19.8 & 43.0 & 81.4\% \\
    EE-MLLM  & 68.2 & 67.3 & 69.1 & 25.9 & 57.5 & 37.0 & 60.1 & 25.1 & 51.3 & 97.2\% \\
    \bottomrule
    \end{tabular}
    }
    \caption{Comparisons with LLaVA and Flamingo under the same settings, including LLM, vision encoder and training data.}
    \label{tab:ablation_llava}
\end{table*}

\section{Ablation Study}
\subsection{Implementation Details}
Following LLaVA-v1.5~\cite{liu2023llava_1p5}, we employ Vicuna-7b-v1.5~\cite{zheng2023vicuna} as our LLM, and CLIP-ViT-L-14 as vision encoder. The training data is consistent with LLaVA-v1.5.
To support high-resolution image input, we replace CLIP-ViT-L-14 with SigLIP~\cite{zhai2023sigmoid_siglip}, which is derived from Idefics2~\cite{laurenccon2024matters_idefics2} and supports high-resolution image input. To compare with cross-attention based methods, we reproduce Flamingo under the similar settings, the implementation details are provided in Appendix~\ref{sec:impl_flamingo}.

\subsection{Comparisons with LLaVA and Flamingo}
In Tab.~\ref{tab:ablation_llava}, we reproduce LLaVA and Flamingo under the same settings with EE-MLLM. We follow LLaVA to employ blip-558k for pre-training and adopt LLaVA-665k for fine-tuning. 
For the resolution of $336\times 336$, we employ CLIP-ViT-L-14 as the vision encoder, ensuring complete alignment with LLaVA-v1.5. 
EE-MLLM demonstrates comparable performance to LLaVA on general benchmarks. 
In terms of fine-grained benchmarks, at $336\times 336$ resolution, EE-MLLM exhibits promising performance on AI2D and ChartQA but slightly underperforms LLaVA on OCRBench and TextVQA. The average score for the $336\times 336$ resolution is 46.1, which is 97.9\% of LLaVA's 47.1. 
And the performance of Flamingo is significantly worse than that of EE-MLLM and LLaVA-v1.5, with an average score of only 39.3, which is 81.7\% of LLaVA's score, this underperformance can be attributed to Flamingo's architecture. Flamingo introduces additional learnable parameters and requires alignment of visual tokens across different LLM layers, necessitating a large volume of training data, for which the BLIP-558K and LLaVA-665K datasets are insufficient.
For the resolution of $672\times 672$, we utilize SigLIP as the vision encoder. As shown in Tab.~\ref{tab:ablation_llava}, EE-MLLM obtains comparable results on AI2D and TextVQA, and the average score maintains 97.2\% of LLaVA's performance. 
Similarly, Flamingo only achieves an average score of 43.0, which can be attributed to inadequate training data.

\begin{table}[t]
    \centering
    \resizebox{\linewidth}{!}{
    \begin{tabular}{ccc ccccc c}
    \toprule
    \multicolumn{3}{c}{\textbf{Aligner Components}} & \multicolumn{6}{c}{\textbf{Evaluation Benchmarks}} \\
    \cmidrule(lr){1-3} \cmidrule(lr){4-9}
    \textbf{V} & \textbf{O} & \textbf{FFN} & \textbf{SeedB} & \textbf{Hall} & \textbf{TextVQA} & \textbf{ChartQA} & \textbf{DocVQA} & \textbf{Avg.} \\
    \midrule
    \rowcolor{gray!10}
    \multicolumn{9}{c}{\textbf{Resolution: $336 \times 336$}} \\
    \midrule
     &  &  & 65.3 & 24.4 & 44.8 & 14.4 & 18.4 & 33.5 \\
     & \checkmark & \checkmark & \textbf{66.3} & \textbf{25.9} & 45.5 & \textbf{16.7} & \textbf{19.2} & \textbf{34.7} \\
    \checkmark &  & \checkmark & 65.4 & 25.3 & 45.4 & 15.6 & 19.1 & 34.2 \\
    \checkmark & \checkmark &  & 65.8 & 25.1 & 43.5 & 14.8 & 17.9 & 33.4\\
    \checkmark & \checkmark & \checkmark & 66.1 & 25.7 & \textbf{46.1} & 15.8 & 18.6 & 34.5 \\
    \midrule
    \rowcolor{gray!10}
    \multicolumn{9}{c}{\textbf{Resolution: $672 \times 672$}} \\
    \midrule
     &  &  & 66.1 & 25.2 & 56.8 & 19.7 & 28.6 & 39.3 \\
     & \checkmark & \checkmark & 66.6 & 25.5 & 58.2 & 23.0 & 29.8 & 40.6 \\
    \checkmark &  & \checkmark & 66.1 & 24.3 & 58.2 & 21.6 & 30.1 & 40.1 \\
    \checkmark & \checkmark &  & 66.4 & 24.1 & 56.3 & 18.5 & 26.5 & 38.4 \\
    \checkmark & \checkmark & \checkmark & \textbf{66.9} & \textbf{27.6} & \textbf{58.8} & \textbf{24.0} & \textbf{31.8} & \textbf{41.8} \\
    \bottomrule
    \end{tabular}
    }
    \caption{Ablation study of the aligner components V, O, and FFN. Checkmarks indicate employed weights. To reduce ablation costs, we adopt SigLIP with dynamic resolution from Idefics2, enabling direct evaluation at $672\times672$ from models trained at $336\times336$.}
    \label{tab:ablation_crossattn_variant}
\end{table}

\begin{figure*}[t]
    \centering
    \includegraphics[width=0.75\linewidth]{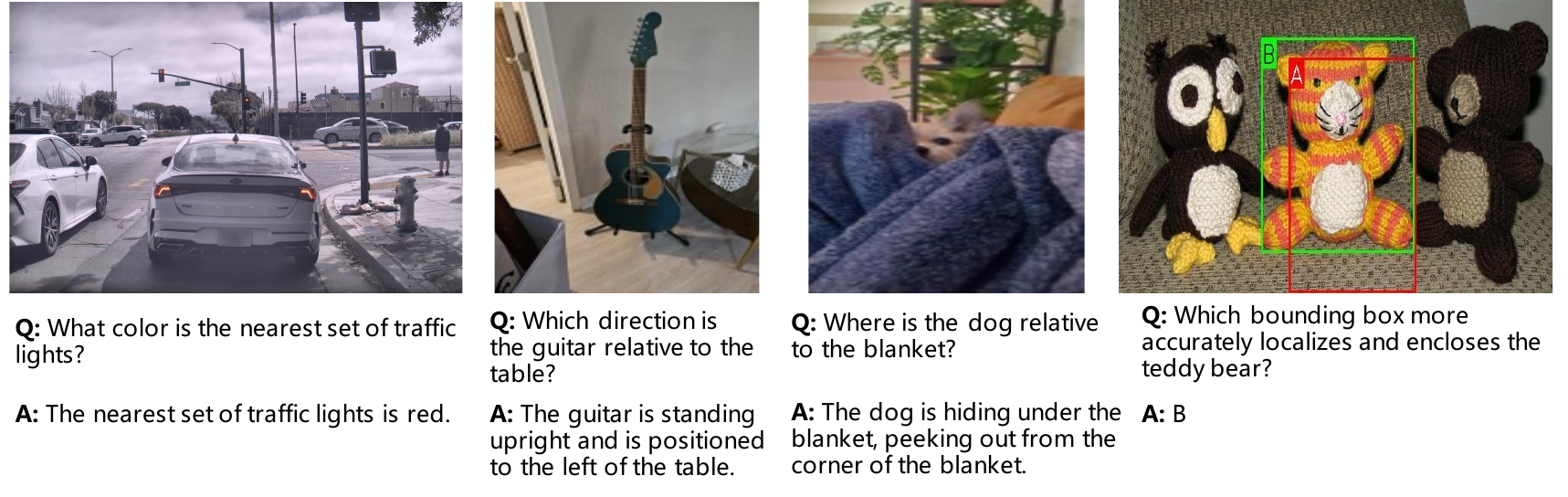}
    \caption{The visualization results of examples sampled from BLINK and RealWorldQA. Answers are generated by our EE-MLLM.}
    \label{fig:visualization}
\end{figure*}

\subsection{Ablation Study of the Aligner}
In Tab.~\ref{tab:ablation_crossattn_variant}, we remove the different weights from the aligner and train the model at a resolution of $336\times 336$. Notably, to reduce the cost of ablation experiments, we adopt SigLIP with dynamic resolution support, derived from Idefics2, as our vision encoder. Thus, models trained at $336\times 336$ can be directly evaluated at a resolution of $672\times 672$.

\noindent We have three findings: 

1) Removing the entire aligner, as depicted in the first row, significantly reduces performance across multiple benchmarks. Specifically, the TextVQA score drops from 46.1 to 44.8, and the average score decreases from 34.5 to 33.5. This result highlights the aligner's effectiveness in aligning visual features with the LLM feature space, allowing text tokens to capture crucial visual information and address questions through causal cross-attention modules.

2) When ablating individual weights in the aligner, we find that maintaining the structure is more critical. The absence of $V$ or $O$ has a relatively minor impact on low-resolution input, even resulting in slightly higher average performance when $V$ is missing. However, when the FFN is missing, the aligner's structure no longer resembles a transformer block, leading to a significant performance loss.

3) The importance of a complete aligner becomes even more pronounced at higher resolution ($672 \times 672$). When $V$ or $O$ weights are removed, we observe substantial performance drops on detail-oriented benchmarks like TextVQA, with scores decreasing from 58.8 to 58.2, ChartQA from 24.0 to 21.6, and DocVQA reducing from 31.8 to 30.1. This demonstrates that all components are essential for effectively processing high-resolution visual inputs, where fine-grained details significantly impact performance.

\subsection{Ablation Study of Training-free Adaption}
In Tab.~\ref{tab:ablation_training_free}, we explore different layer configurations for training-free adaptation of EE-MLLM. Using LLaVA-v1.5-7b with 32 layers as our baseline, we investigate various configurations of attention mechanism modification.

The first configuration represents the original LLaVA, where all layers from 0 to 31 use the standard self-attention mechanism, achieving an average score of 35.5. When we apply the composite attention mechanism to layers 8 to 31, we observe a slight performance decrease to 32.4 on average, with improvements only on HallusionBench.

Notably, the third configuration, which employs composite attention for layers 16 through 31 while maintaining self-attention for layers 0 through 15, achieves the best overall performance with a 36.5 average. This configuration not only maintains comparable performance on SeedBench, TextVQA, ChartQA, and DocVQA compared to the baseline, but also demonstrates significant improvement on HallucinationBench with a score of 32.4 versus 25.7.

To further validate the effectiveness of our composite attention mechanism, we implement a configuration in the fourth row that maintains self-attention only in layers 0 through 15 and drops visual tokens in deeper layers. This results in a substantial performance degradation across fine-grained benchmarks including TextVQA, ChartQA, and DocVQA, with the average score dropping to 27.2. This finding confirms the composite attention mechanism plays a crucial role in preserving performance on tasks requiring detailed visual understanding, while simply dropping visual information in deeper layers is detrimental.

These experimental findings clearly suggest that applying composite attention selectively to deeper layers from 16 to 31 provides an optimal balance between efficiency and overall performance across diverse multimodal tasks.

\begin{table}[t]
    \centering
    \resizebox{\linewidth}{!}{%
    \begin{tabular}{c@{\hspace{8pt}}c@{\hspace{8pt}}c@{\hspace{8pt}}c@{\hspace{8pt}}c@{\hspace{8pt}}c@{\hspace{8pt}}c@{\hspace{8pt}}c}
        \toprule
        \multicolumn{2}{c}{\textbf{Layer Configuration}} & \multicolumn{6}{c}{\textbf{Evaluation Benchmarks}} \\
        \cmidrule(lr){1-2} \cmidrule(lr){3-8}
        \makecell{\textbf{Self-attn}\\ \textbf{Layers}} & \makecell{\textbf{Comp-attn}\\ \textbf{Layers}} & \textbf{SeedB} & \textbf{Hall} & \textbf{TextVQA} & \textbf{ChartQA} & \textbf{DocVQA} & \textbf{Avg.} \\
        \midrule
        $[0,31]$ & ---  & 65.1  & 25.7 & 46.1 & 17.9 & 22.5 & 35.5  \\
        $[0,7]$  & $[8,31]$  & 62.8  & 29.0 & 39.8 & 16.1 & 14.5 & 32.4  \\
        $[0,15]$ & $[16,31]$ & 65.1  & 32.4 & 45.5 & 17.4 & 21.9 & 36.5  \\
        $[0,15]$ & --- & 65.1  & 29.2 & 17.6 & 13.8 & 10.5 & 27.2  \\
        \bottomrule
    \end{tabular}%
    }
    \caption{
    Ablation study on layer configuration for training-free adaptation. ``Self-attn Layers" indicates which layers maintain the original self-attention mechanism, while ``Comp-attn Layers" shows which layers are modified to use the composite attention mechanism. The notation $[i, j]$ indicates layers from index $i$ to $j$.
    }
    \label{tab:ablation_training_free}
\end{table}

\subsection{Visualization}
We present four examples sampled from BLINK~\cite{fu2024blink} and RealWorldQA~\cite{realworldqa} to assess the impact of the architectural changes in Fig~\ref{fig:visualization}. 
The first example demonstrates that EE-MLLM can perceive fine-grained visual context within the image, such as the color of traffic lights. The second and third examples highlight EE-MLLM's ability to comprehend the position of objects. Specifically, EE-MLLM can accurately identify guitar's position.

%% file: Sections/5_conclusion.tex
\section{Conclusion}
In this paper, we introduce EE-MLLM, a novel framework that unifies data efficiency and compute efficiency through two key innovations: (1) a composite attention mechanism that eliminates redundant self-attention computations among visual tokens, reducing FLOPs by 24\% compared to self-attention methods, and (2) a parameter-free aligner that reuses existing LLM weights to align visual and text modalities, minimizing training data requirements.
EE-MLLM achieves state-of-the-art performance across various benchmarks. Crucially, it reduces inference latency from 277 ms to 79 ms, enabling real-time applications such as content moderation and quality control. The training-free variant (EE-MLLM-F) further demonstrates adaptability, maintaining performance while cutting computational costs by selectively applying composite attention to deeper layers.

%% file: Sections/6_appendix.tex
\clearpage
\setcounter{page}{1}
\twocolumn[{
    \renewcommand\twocolumn[1][]{#1}
    \maketitlesupplementary
    \begin{center}
        \includegraphics[width=1.0\linewidth]{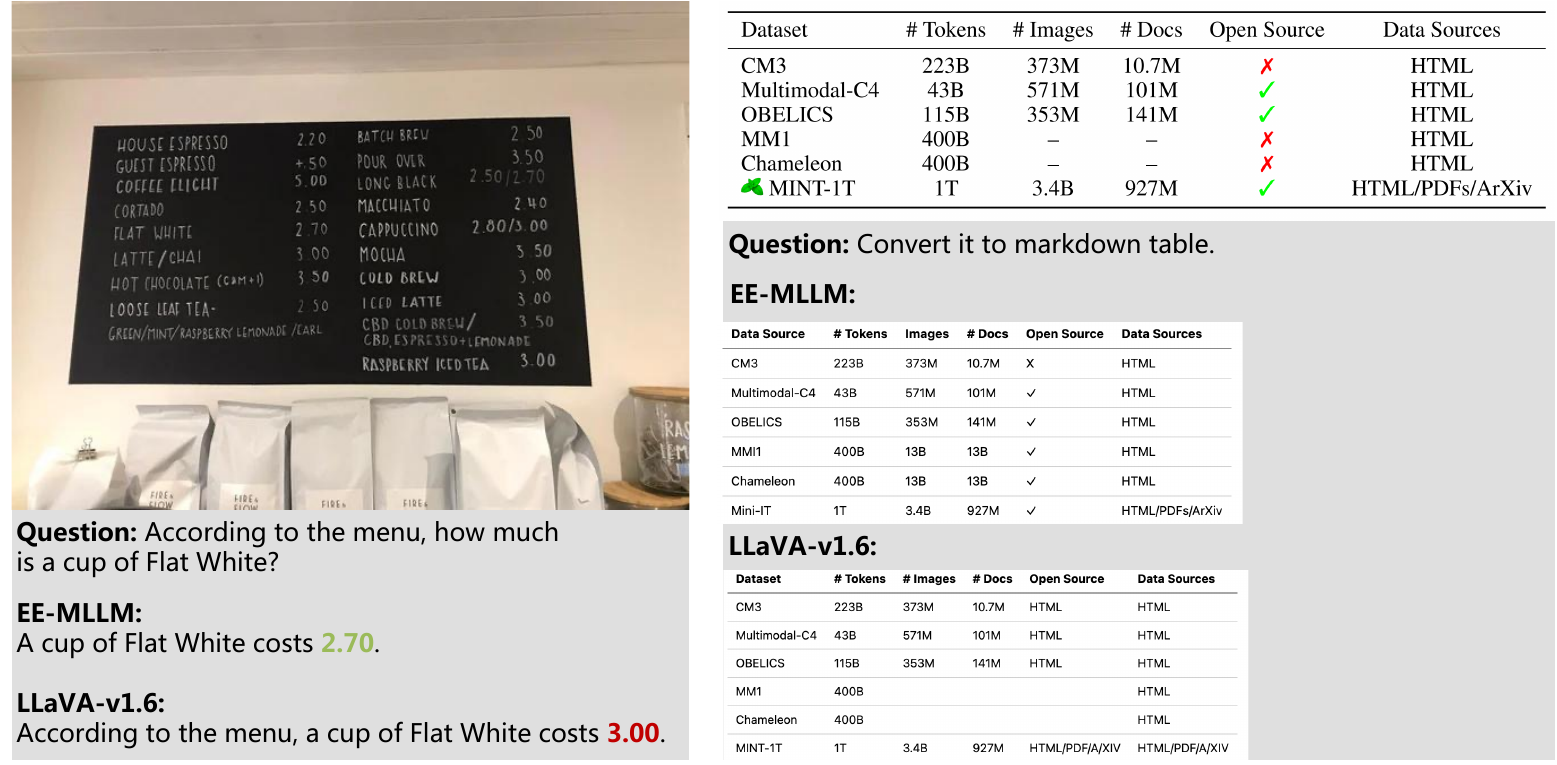}
        \captionof{figure}{Comparison of EE-MLLM and LLaVA-v1.6 response results on challenging samples.}
        \label{fig:appendix_1}
    \end{center}
}]

\section{Visualization}
Fig.~\ref{fig:appendix_1} provides two examples comparing EE-MLLM with LLaVA-v1.6~\cite{liu2024llavanext}. The first image, a handwritten coffee menu, prompts a question about the image's fine-grained details. Our EE-MLLM successfully identifies the price of the Flat White listed in the image, providing the correct answer, whereas LLaVA-v1.6 does not. The second example requires the model to convert the image's table into markdown format. Our observations indicate that EE-MLLM's response is more accurate, while LLaVA makes mistakes in the columns labeled ``Open Source'' and ``Data Source''.

\section{Implementation Details}
\subsection{Supervised Fine-tuning Data}
\label{sec:sft_data}
The details of our supervised fine-tuning data are shown in Tab.~\ref{tab:sft_data}
Following Deepseek-VL~\cite{lu2024deepseek}, our supervised fine-tuning data incorporates open-source gpt4v datasets, including ShareGPT4V~\cite{chen2023sharegpt4v}, LVIS-Instruct4V~\cite{wang2023see_lvis_instruct4v}, LAION-GPT4V~\cite{laion-gpt4v}, and TextOCR-GPT4V~\cite{textocr-gpt4v}. Moreover, to enhance the fine-grained capabilities of MLLM, we introduce DVQA~\cite{kafle2018dvqa}, PlotQA~\cite{methani2020plotqa}, CLEVR~\cite{johnson2017clevr}, DocVQA~\cite{mathew2021docvqa}, ChartQA~\cite{masry2022chartqa}, and ScienceQA~\cite{lu2022learn_scienceqa} for fine-tuning. To preserve language abilities during fine-tuning, we utilize a large volume of text-only data from various sources. 
The text-only data encompasses various dataset types, such as mathematical reasoning, multi-round dialogues, logic comprehension, and code. Meanwhile, the in-house data primarily consists of OCR-related, table and diagram comprehension, and mathematical reasoning data.

\subsection{Implementation of Flamingo}
\label{sec:impl_flamingo}
In the ablation study, we reproduced Flamingo. Like LLaVA-v1.5-7B, we used CLIP-ViT-L-14 as the vision encoder and Vicuna-7B as the LLM. We introduced cross-attention blocks every 4 layers into the LLM for inter-modal interaction. The cross-attention block includes attention operations and MLP. Specifically, we inserted cross-attention blocks at layers 3, 7, 11, 15, 19, 23, 27, and 31. The training setup was consistent with LLaVA, including training data, learning rate, and optimizer. In the ablation, we ensured a fair comparison.

\begin{table}[t]
    \centering
    \resizebox{0.8\linewidth}{!}{
    \begin{tabular}{>{\raggedright\arraybackslash}p{3.5cm}|>{\raggedright\arraybackslash}p{5cm}|c}
        \toprule
        \textbf{Type} & \textbf{Dataset} & \textbf{Ratio} \\
        \midrule
        \multirow{6}{*}{General Instruction Data} 
            & ShareGPT4V & \multirow{6}{*}{35\%} \\
            & LVIS-Instruct4V & \\
            & LAION-GPT4V & \\
            & TextOCR-GPT4V & \\
            & Localized Narratives & \\
            & IconQA & \\
        \midrule
        \multirow{6}{*}{Text-Centric Data} 
            & DVQA & \multirow{6}{*}{8.5\%} \\
            & PlotQA & \\
            & CLEVR & \\
            & DocVQA & \\
            & ChartQA & \\
            & ScienceQA & \\
        \midrule
        \multirow{2}{*}{Others} 
            & Text-only data & 47\% \\
            & In-house data & 9.5\% \\
        \bottomrule
    \end{tabular}
    }
    \caption{Details of supervised fine-tuning data.}
    \label{tab:sft_data}
    
\end{table}

\section{Evalution Benchmarks}
We conduct our evaluations using the VLMEvalKit~\cite{duan2024vlmevalkit}, and the results of other state-of-the-art models are obtained from the same source.
\subsection{General Benchmarks}
\label{sec:general_benchmarks}
1) MMBench-EN~\cite{MMBench} is a comprehensive multimodal benchmark specifically designed to assess the performance of MLLMs. It comprises over 3,000 multiple-choice questions spanning 20 ability categories. We evaluate EE-MLLM on the MMBench-EN-V1.1.
2) MME~\cite{fu2023mme} evaluates advanced MLLMs in terms of perception and cognition, encompassing a total of 14 subtasks. To minimize the impact of prompt engineering on MLLMs, MME's instructions are designed to elicit simple binary responses, such as ``please answer yes or no.". We report the results of the perception split of MME.
3) ScienceQA~\cite{lu2022learn_scienceqa}is derived from elementary and high school science curricula. The questions in ScienceQA cover three subjects: natural science, language science, and social science.
4) HallusionBench~\cite{guan2023hallusionbench} is designed for evaluating image-context reasoning, comprising 346 images paired with 1,129 questions crafted by human experts. HallusionBench takes into account both language hallucinations and visual illusions across a diverse range of topics.
5) MMMU~\cite{yue2023mmmu} collects 11.5K multimodal questions from college exams, quizzes, and textbooks, covering six core disciplines, spanning 30 subjects and 183 subfields, including 30 heterogeneous image types.
6) CCBench~\cite{MMBench}, developed by the MMBench team, is specifically designed to evaluate MLLMs in the domain of Chinese Culture.
7) SeedBench~\cite{li2023seed} encompasses 19K multiple choice questions, covering 12 evaluation dimensions, including both image and video. We only use questions with images for evaluation.
8) BLINK~\cite{fu2024blink} contains 14 visual perception tasks which pose significant challenges for current
multimodal LLMs. 

\begin{figure}[t]
    \centering
    \includegraphics[width=1.0\linewidth]{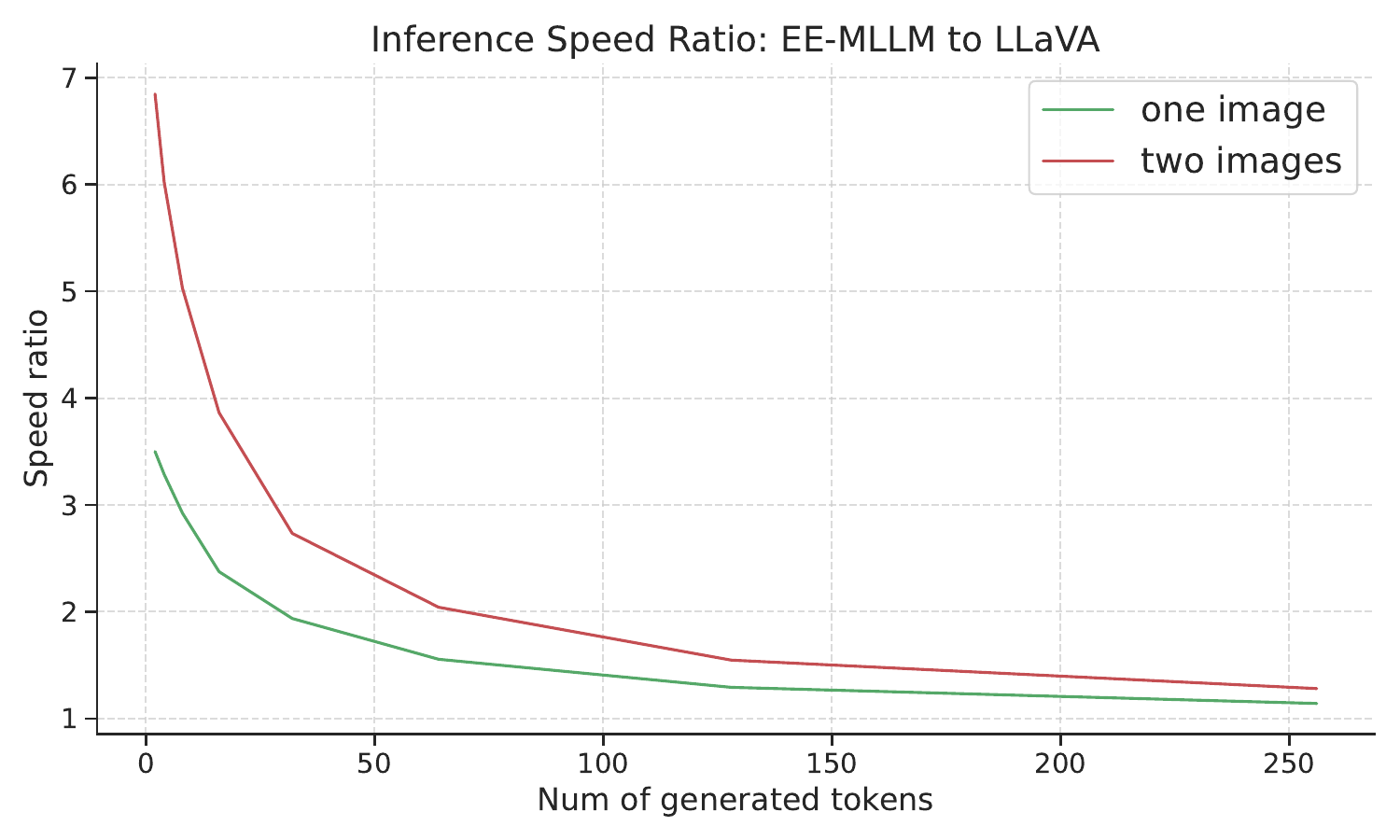}
    \caption{Inference speed ratios of EE-MLLM and LLaVA for single and double image inputs.}
    \label{fig:multi_img_inference}
\end{figure}

\subsection{Fine-grained Benchmarks}
\label{sec:finegrained_benchmarks}
1) AI2D~\cite{kembhavi2016diagram_ai2d} emphasizes diagram interpretation and reasoning, comprising 5,000 diagrams and 15,000 questions and answers.
2) OCRBench~\cite{liu2023hidden_ocrbench}
aims to facilitate the assessment of MLLM OCR capabilities, including 29 datasets.
3) TextVQA~\cite{singh2019textvqa}
consists of 45,336 questions and 28,408 images that necessitate reasoning about text for answering. We use the validation set, containing 5,000 images, for evaluation.
4) ChartQA~\cite{masry2022chartqa} is a large-scale benchmark featuring 20,882 charts, with questions focusing on logical and visual reasoning.
5) DocVQA~\cite{mathew2021docvqa}
concentrates on document image understanding, encompassing 50,000 questions and over 12,000 images. We evaluate using the validation set, which includes 5,349 questions and 1,286 images.
6) Seed2 Plus~\cite{li2024seedbench_plus} specifically designed for text-rich visual comprehension evaluation of MLLMs, includes 2.3K multiple-choice questions covering charts, maps, and webs.

\section{Comparison of Overall Inference Speed}
\label{sec:comp_overall_inference_speed}
For MLLMs, the overall inference time consists of the prefilling time and the decode time. The prefilling time mainly involves processing visual inputs and textual instructions, while the decode time refers to the duration required to generate new tokens.

In the main text, we have demonstrated EE-MLLM's substantial improvements over LLaVA in reducing prefilling time. Here, we further compare the overall inference speed. In Fig.~\ref{fig:multi_img_inference}, we present the speed ratio between EE-MLLM and LLaVA when generating varying numbers of tokens.
We conduct the comparisons on a single NVIDIA H800. The resolution of the input image is set to 980×980, and the number of generated tokens varies from 2 to 256. 

Our findings show that when generating 8 tokens, EE-MLLM’s inference speed is three times that of LLaVA. However, as the number of generated tokens increases, the speed ratio decreases. When generating 64 tokens, EE-MLLM’s inference speed is 1.6 times that of LLaVA.  The reason for this observation is that our EE-MLLM primarily reduces the computational cost during the prefilling stage, which calculates the KV Cache of visual tokens. The generation of the first token is faster than self-attention-based methods like LLaVA. However, the advantage in inference speed diminishes after the first token. Although the inference speed of EE-MLLM decreases as the number of generated tokens increases, it still maintains an advantage over LLaVA, especially under multi-image input scenarios.